\documentclass[letterpaper, 10 pt, conference]{ieeeconf}  

\IEEEoverridecommandlockouts                              

\usepackage{times}

\usepackage{multicol}
\usepackage[bookmarks=true]{hyperref}
\usepackage{graphicx}
\usepackage{subcaption}
\graphicspath{{figures/}} 

\usepackage{amsmath}

\title{\LARGE \bf
Learning Social Navigation from Demonstrations with Deep Neural Networks
}

\author{Yigit Yildirim$^{1}$ and Emre Ugur$^{2}$
\thanks{$^{1}$Yigit Yildirim is with Computer Engineering Department, Bogazici University, Istanbul, Turkey
        {\tt\small yigit.yildirim@boun.edu.tr}}%
\thanks{$^{2}$Emre Ugur is with Computer Engineering Department, Bogazici University, Istanbul, Turkey
        {\tt\small emre.ugur@boun.edu.tr}}%
}

\begin{document}

\maketitle
\thispagestyle{empty}
\pagestyle{empty}

\begin{abstract}

Traditional path planning techniques treat humans as obstacles. This has changed since robots started to enter human environments. On modern robots, social navigation has become an important aspect of navigation systems. To use learning-based techniques to achieve social navigation, a powerful framework that is capable of representing complex functions with as few data as possible is required.
In this study, we benefited from recent advances in deep learning at both global and local planning levels to achieve human-aware navigation on a simulated robot. Two distinct deep models are trained with respective objectives: one for global planning and one for local planning. These models are then employed in the simulated robot. In the end, it has been shown that our model can successfully carry out both global and local planning tasks. We have shown that our system could generate paths that successfully reach targets while avoiding obstacles with better performance compared to feed-forward neural networks.

\end{abstract}

\section{INTRODUCTION}

Mobile robot navigation has been studied for decades. Many notable techniques have been proposed in this area over the years, \cite{burgard1999ExperiencesWA,thrun2000probabilistic, mobot}. These approaches have prioritized the safety and the robustness features, i.e. the principal driving factor behind the development in this field has been the collision avoidance \cite{dwa}. On the other hand, as humans start to share their environments with robots, new requirements for mobile robot navigation have emerged. 

In \cite{nonaka2004evaluation}, physical and mental aspects of the safety are separately evaluated. This separation reveals the need to question the psychological efficiency of navigation systems of mobile robots. Keeping in mind the assumption that humans prefer to interact with machines in the same way that they interact with other people, in order to achieve a natural integration to the environments populated by people, mobile robots must be developed to be not only safe but also comprehensible.


Broadly speaking, human-aware navigation corresponds to the navigation that complies with the social rules of the people. In their own environments, humans tend to work cooperatively to realize social navigation. Then, it is only natural to imitate this behavior on the robots to achieve socially-acceptable navigation. However, imitating people introduces new constraints to be satisfied by the navigation systems of robots.

These constraints have been addressed in many studies in the literature. Essentially, these studies can be divided into two categories: manually-encoded controllers and learning-based ones. One of the notable studies of the first category is the Social Force Model (SFM) \cite{helbing1995social}. Based on the behavioral techniques from social sciences, SFM suggests that pedestrians move under the effect of certain abstract forces, just like the particles in an electrical field. While the navigational goal attracts the pedestrian, obstacles and other people exert repulsive forces. Despite its wide application\cite{zanlungo2011social, ferrer2013robot, farina2017walking}, some researchers state that not being based on the statistical data is a weakness of the model \cite{kretzschmar2016socially}. 

To create statistics-based socially compliant navigation frameworks, a large number of machine learning algorithms have been employed. One of the popular algorithms is Inverse Reinforcement Learning (IRL) \cite{kitani2012activity,vasquez2014inverse,kim2016socially}. Given the perfect expert demonstrations, IRL tries to identify the underlying reward structure, which in turn can be used by any Reinforcement Learning (RL) algorithm to create a human-aware navigation policy. Even though the justification of the unfixed reward function is appealing, the features that shape the reward function are assumed to be known, which is considered as a strong assumption \cite{wulfmeier2015maximum}. Generally in this domain, feature engineering leads to strong assumptions. This problem can be solved by extracting the social behaviors and navigation strategies of pedestrians directly from the data. This is challenging because the controller needs to be complex enough to capture the non-linearities in the data.

To address this issue in social navigation domain, deep learning techniques have been used. In \cite{chen2017}, Deep Reinforcement Learning is used to obtain a socially plausible navigation policy. As in other RL approaches, this procedure relies on a predefined reward which is difficult to obtain. Imitation Learning skips the reward extraction and tries to learn policies directly from the data. In \cite{tai2018socially} and \cite{gupta2018social}, Generative Adversarial Networks are used for this purpose. These approaches are complex enough to overcome the aforementioned issues. However, these models need too much data to be trained \cite{che2020efficient}. On the other hand, the preferred system needs to learn from a small dataset and to generalize to novel configurations.

Moreover, the majority of the studies on this domain target only the local controller of the robot as it is the part that creates motion commands to drive the robot. However, using only the local controller makes the robot vulnerable to local minima \cite{koren}. Today, typical robotic navigation systems adopt the two-layered hierarchical approach for path planning tasks. Given a map of the environment, a robot firstly calculates a trajectory in the so-called \textit{global planning} phase. Then, the robot follows the computed trajectory with a controller in the so-called \textit{local planning} phase.

In this paper, we use Conditional Neural Processes (CNPs) \cite{pmlr-v80-garnelo18a} in order to address the issues mentioned above in both global and local planning phases. CNPs can be modified to generate complete trajectories to replace the global planner. Also, they can create goal-directed behavior while actively avoiding obstacles. This characteristic makes it a candidate for the local planner, as well. CNPs extract the prior knowledge directly from the training data by sampling observations from it, and uses it to predict a conditional distribution over any other target points. CNPs can learn complex temporal relations in connection with external parameters and goals. In this paper, we present the initial results of our system. Upon successful preliminary results with this conceptual model, we aim to extend this work to integrate our path planning system into an actual robot in another study.

\section{Related Work}
Traditionally, approaches to solve the path planning problem can be divided into two categories based on the environmental knowledge they use: deliberate and reactive. Deliberate planners exploit the environmental knowledge by means of static maps and calculate the robot's trajectory before execution. On the other hand, reactive planners rely on sensory information to deal with local parts of the environment. Either approach has its advantages and drawbacks. Hence, the evolution of the path planning approaches leads to the combination of these two approaches. Hybrid frameworks have been the typical approach for many years, as explained in \cite{MEYER2003283}.

In the following, we elaborate on this conventional framework's building blocks and the social navigation concept.

\subsection{Hierarchical Path Planning}
The standard hybrid path planning framework combines the strengths of deliberate and reactive planners. It consists of a two-phased procedure in a hierarchical manner; global planning is for the deliberation and local planning for the reactivity.

\subsubsection{Global Path Planner}
In the first phase of a standard hierarchical path planning pipeline, a global planning procedure is applied. On the static map of the environment, the function of a global planner is to generate a path from the starting position to the destination. Conventionally, many graph search algorithms have been applied to calculate the trajectory between initial and goal configurations, the most popular being A* explained in \cite{astar}. For a more complete list of global planning approaches, see \cite{giesbrecht2004global}.

The global planning itself is not sufficient to navigate the robot between two points. Local planning is needed to create velocity commands that handle the cases with new or dynamic obstacles.

\subsubsection{Local Path Planner}
In order to realize computed trajectories, the local planning procedures are used in the second phase of hierarchical path planning. The most prominent objective of the local planner is to generate velocity commands so that the robot can follow the computed trajectory. In addition, by using the sensory information about the robot's surroundings, it is the local planner's duty to avoid obstacles. There are many local planning algorithms in the literature, such as \cite{vfh, dwa, apf, teb, zhu2006robot, vadakkepat2000evolutionary}. For a more complete list, see \cite{cai2020mobile}.

On the other hand, despite being quite safe, these traditional controllers take no account of social norms. They consider people as obstacles to be avoided. Recent attempts to create local controllers that consider these norms has paved the way for social robot navigation.

\subsection{Social Navigation}
According to \cite{kruse2013human}, the benefits of social navigation are threefold: it increases the comfort of the people around the robot, it improves the naturalness of the robotic platform and it also enhances the sociability of the robot. Furthermore, in \cite{nonaka2004evaluation}, physical and mental aspects of safety are separately evaluated. For us, this separation reveals the need to question the psychological efficiency of navigation systems of mobile robots.

\begin{figure}[h]
\begin{center}
\includegraphics[width=0.8\columnwidth]{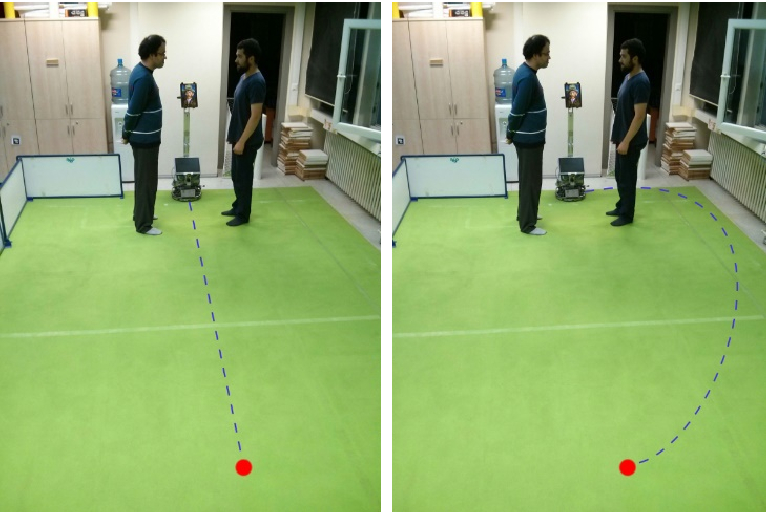}
\end{center}
\caption{Comparison between regular and social navigation.}
\label{fig:sn}
\end{figure}

The concept of social navigation lies in the intersection of two concepts: navigation and human-robot interaction. It describes improving the navigation of the robot to enhance its comprehensibility by the humans around.  Figure \ref{fig:sn} is rather self explanatory. On the left, we see a robot with a perfectly safe navigation plan. In contrast, although non-optimal, the planned path on the right is socially compliant.

\begin{figure*}
\begin{center}
\includegraphics[width=\textwidth]{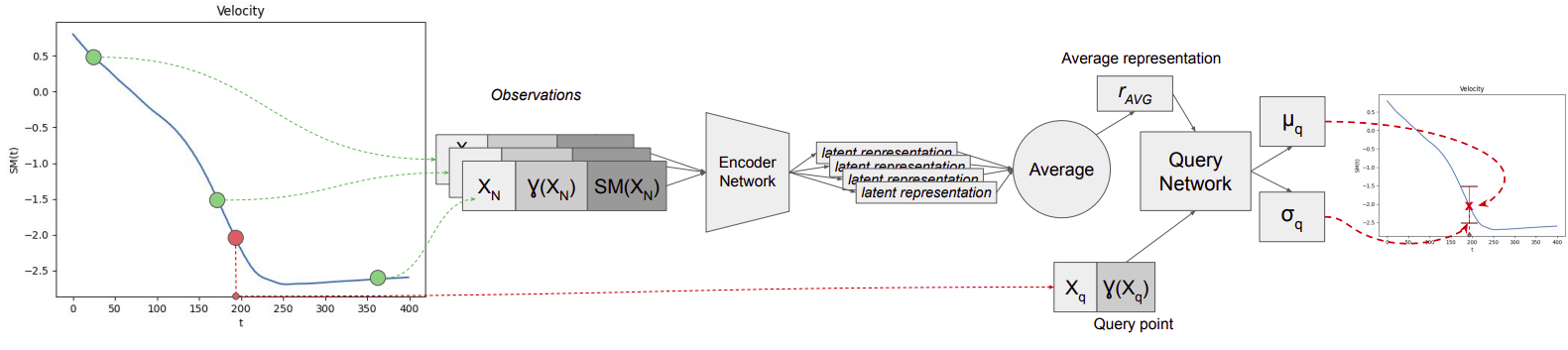}
\caption{General layout of the training phase of our model.}
\label{fig:cnmp}
\end{center}
\end{figure*}

\section{Method}
In this work, we address two parts of the hierarchical path planning individually and show that the capabilities of the model we propose can handle both global and local planning. We suggest employing a variant of Conditional Neural Processes (CNPs) for both of them separately. 

CNP is a powerful deep learning framework, which is inspired by the flexibility of stochastic processes, but organized as neural networks and trained with gradient descent \cite{pmlr-v80-garnelo18a}. Since its emergence, CNPs and variants have been successfully applied in several robot learning problems \cite{seker2019conditional, yin2019meta, akbulut2020adaptive}. Instead of outputting a single value, CNP learns a Gaussian distribution over the demonstrated trajectories. The set $ D $, representing all demonstrations is defined as follows: $D = \{D_i\}_{i=0}^N$, where each $D_i$ is a trajectory of a number of points in a high-dimensional space. Essentially, $D_i=(X_t, \gamma(X_t), SM(X_t))_{t=0}^{\tau}$, where $X$ is the state variable, $\gamma(X)$ is a function representing task parameters and $SM(X)$ is the sensorimotor function to be learned. The encoder network produces a latent representation for each trajectory and these representations are passed through an averaging operation to create a compact representation $r_{AVG}$ for the task at hand. Subsequently, $X_q$, $\gamma(X_q)$ and $r_{AVG}$ are fed to the \textit{Query Network} to produce an estimate for $SM(X_q)$. $\mu_q$ and $\sigma_q$ respectively represent the estimated mean and the variance. Figure \ref{fig:cnmp} shows the overall model.


The model consists of an encoder network which outputs latent representations by using the sampled points on the demonstrated trajectories. These representations in the latent space are then averaged to come up with a compact representation of the trajectory. At query time, this compact representation is concatenated with the target point and the resulting vector is fed to the query network to generate the estimated sensorimotor response of the model. 

\subsubsection{Global Planning}
One of the most powerful aspects of the CNPs approach is its ability to generate complete trajectories. Upon training the encoder and query networks, target points can be simultaneously processed from the starting point to the end to create an entire trajectory. This ability can be exploited to create global plans in the first phase of a hierarchical path planning procedure.

\subsubsection{Local Planning}
We also benefit from CNPs in reactively responding to the changes in its domain. With this, we substitute the local planning module of the hierarchical path planners with local CNPs. This requires sensory input to be processed by the CNP as task parameters. In the current study, high-level parameters such as distance to the obstacles or relative position to the goal point are used as input to the local CNPs. It was shown that CNPs can efficiently handle low-level and high-dimensional input as well, as shown in \cite{gordon2019convolutional}.


\section{Experiments and Results}

\subsection{Environment}
\begin{figure}[h]
  \begin{center}
    \begin{subfigure}{0.49\columnwidth}
        \includegraphics[width=\linewidth]{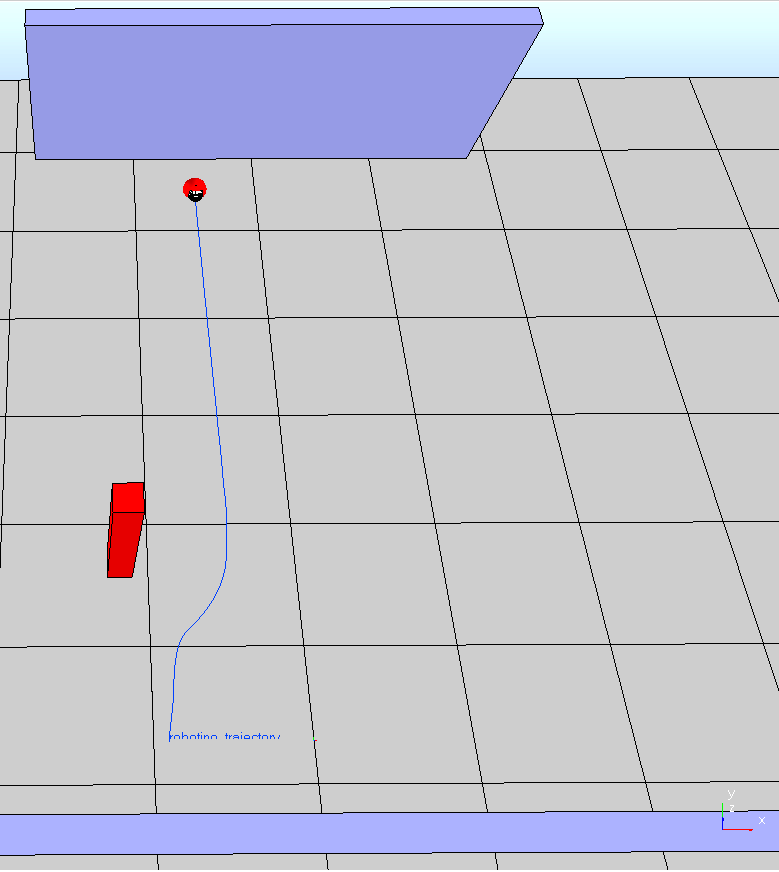}
    \end{subfigure}
    \hfill
    \begin{subfigure}{0.49\columnwidth}
        \includegraphics[width=\linewidth]{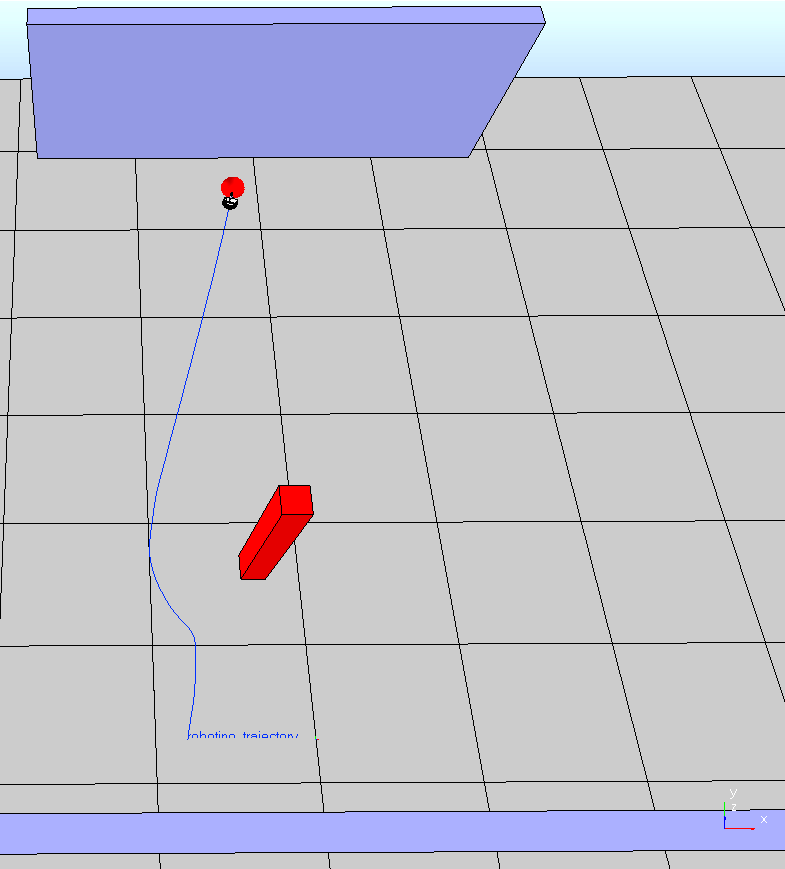}
    \end{subfigure}
    \caption{Data collection on the simulation. The motion trajectory is shown with blue line.}
    \label{fig:dc}
  \end{center}
\end{figure}

Our system was verified in CoppeliaSim simulation environment \cite{coppeliaSim} that includes an omnidirectional robot platform (Robotino \cite{robotino}). 
The Social Force Model, described in \cite{helbing1995social}, is implemented as the local controller of the robot to gather demonstration trajectories. With the assumption that it generates socially plausible trajectories, 1000 trajectories with randomly different starting, goal and obstacle poses are recorded. Single, multiple, stationary and dynamic objects are placed at random positions in each trial. The data collection process is shown in Figure \ref{fig:dc}.

\subsection{Global Planner}

\begin{figure}[h]
  \begin{center}
    \begin{subfigure}{\columnwidth}
        \includegraphics[width=\linewidth]{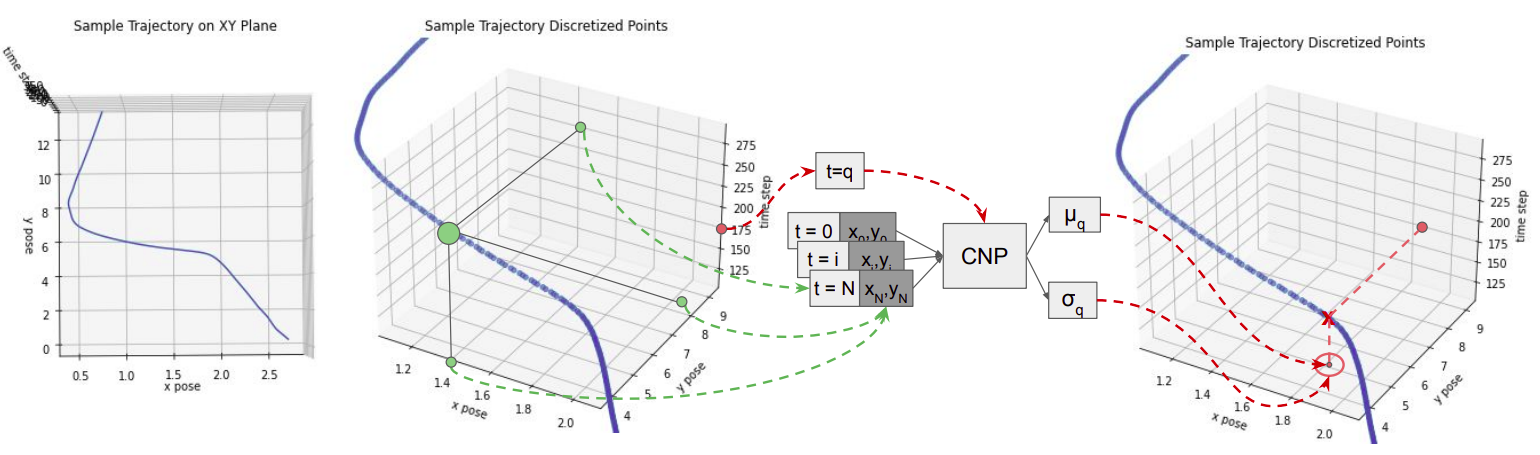}
        \caption{Training the network with a randomly chosen demonstration trajectory.}
        \label{fig:cnmp_global_tr}
    \end{subfigure}
    \vskip\baselineskip
    \begin{subfigure}{\columnwidth}
        \includegraphics[width=\linewidth]{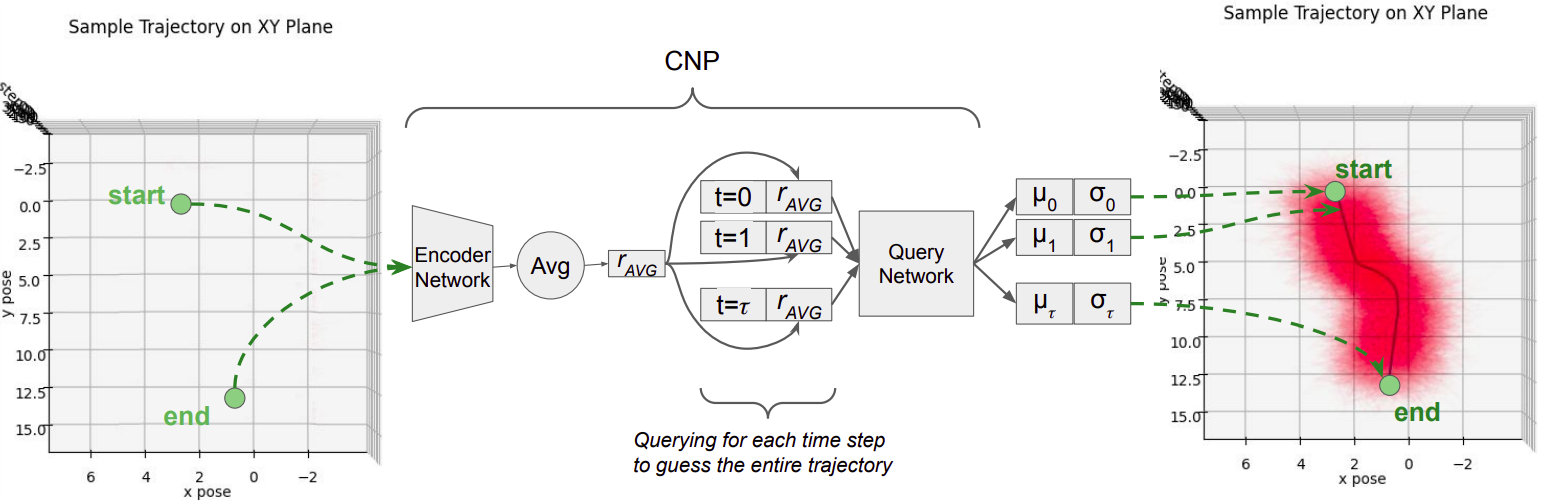}
        \caption{Generating a global path in test phase.}
        \label{fig:cnmp_global_test}
    \end{subfigure}
    \caption{CNP as the global planner.}
  \end{center}
\end{figure}

To show the path planning capability of our method, the model is fed with the entire trajectories of positions of the robot and trained on these demonstrations. The representation of the data is as follows:
\begin{align*}
&X=time\_step \\
&\gamma(X)=(start\_x, start\_y, goal\_x, goal\_y, obs\_x, obs\_y) \\
&SM(X)=(position\_x, position\_y),
\end{align*}
where $obs\_x$ and $obs\_y$ refers to the obstacle's $x$ and $y$ positions. Fig.~\ref{fig:cnmp_global_tr} illustrates the training phase and Fig.~\ref{fig:cnmp_global_test} shows how the entire path is queried. 

\begin{figure}
\begin{center}
\includegraphics[width=\columnwidth]{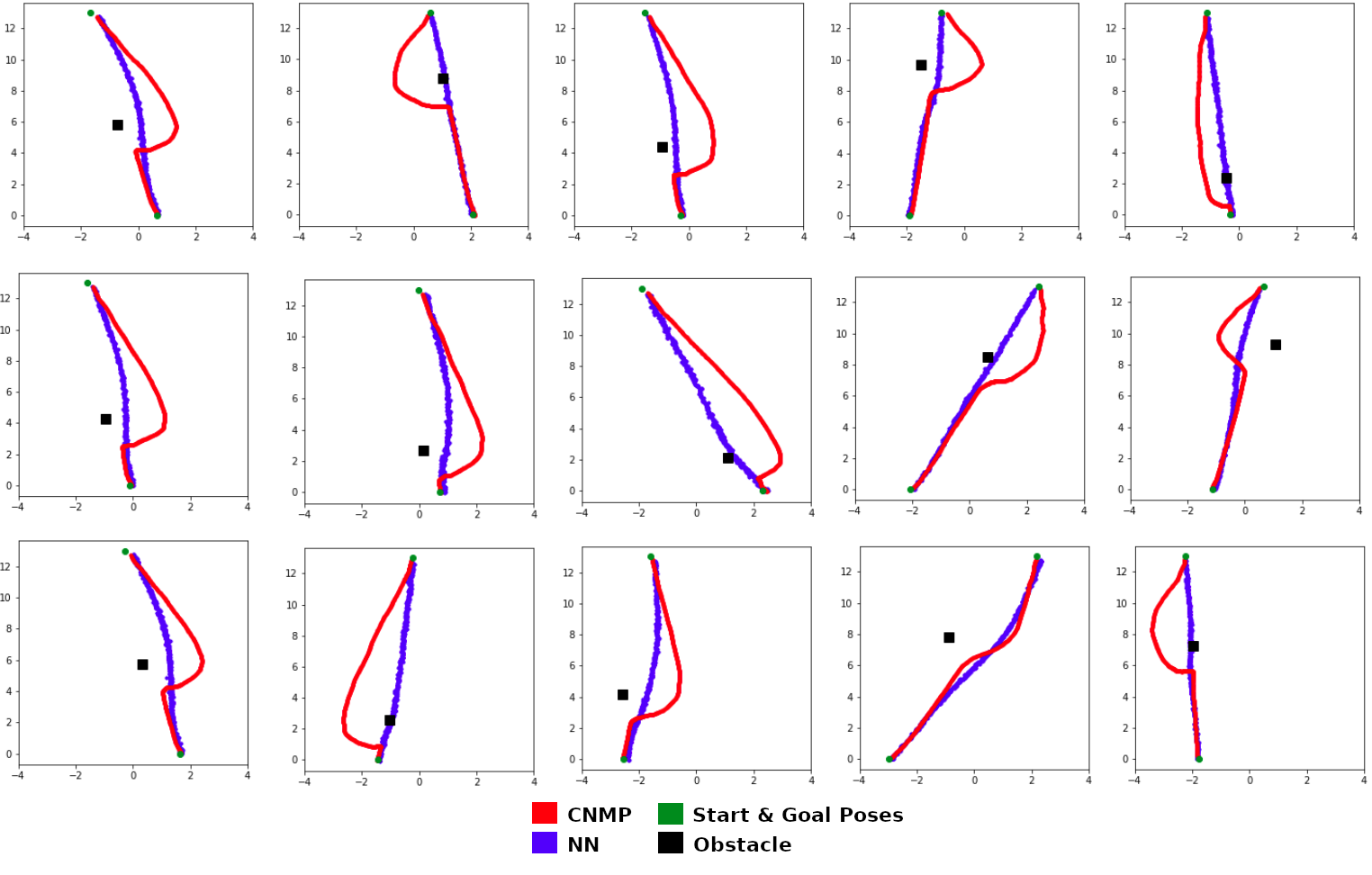}
\end{center}
\caption{Comparison between our global planner network (CNP) and a 5-layered feed-forward neural network (NN) on global planning in sample environments.}
\label{fig:cnmp_nn}
\end{figure}

To show the strength of CNPs over standard neural networks, we compare their performance on the trajectory planning task. For this purpose, we implemented a 5-layered standard feed-forward neural network and trained it on the same dataset of 1000 trajectories. The comparison of their performances on a global planning task is given in Figure \ref{fig:cnmp_nn}. This result shows that while a feed-forward neural network cannot generate global paths that avoid obstacles, our system can. We believe that this is due to the capability of our system to learn multiple-modes of operations. Standard feed-forward networks, given demonstration paths that avoid obstacles from different sides, probably interpolates these paths; whereas our system can learn to generate trajectories from both sides.

\subsection{Local Planner}
From the local perspective, the input parameters of our local network are distance-to-goal, distance-to-obstacle and velocity commands. Here, the formulation of the problem is as follows:
\begin{align*}
&X=(distance\_to\_goal\_x, distance\_to\_goal\_y) \\
&\gamma(X)=(distance\_to\_obs\_x, distance\_to\_obs\_y) \\
&SM(X)=(velocity\_x, velocity\_y)
\end{align*}

Note this time that, we do not use a linearly increasing phase variable, as we did in the case of global planning. Conditioned on the starting and destination poses, the use of the task parameter $\gamma(X)$ gave the model the ability to reactively change the velocity commands with respect to changing obstacle positions.

\begin{figure}[h]
\begin{center}
\includegraphics[width=\columnwidth]{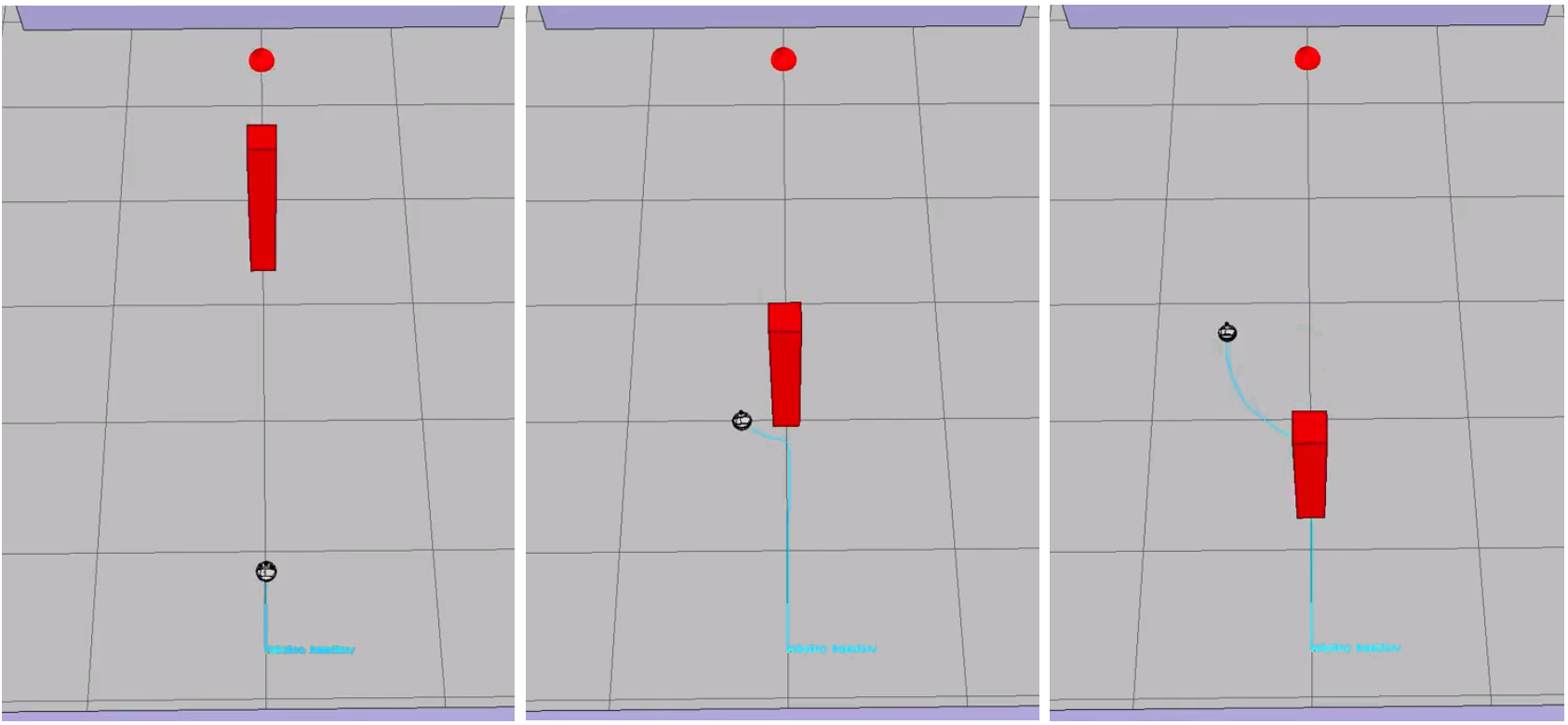}
\end{center}
\caption{Robot is avoiding from a vertically moving obstacle.}
\label{fig:vert}
\end{figure}


\begin{figure}[h]
\begin{center}
\includegraphics[width=\columnwidth]{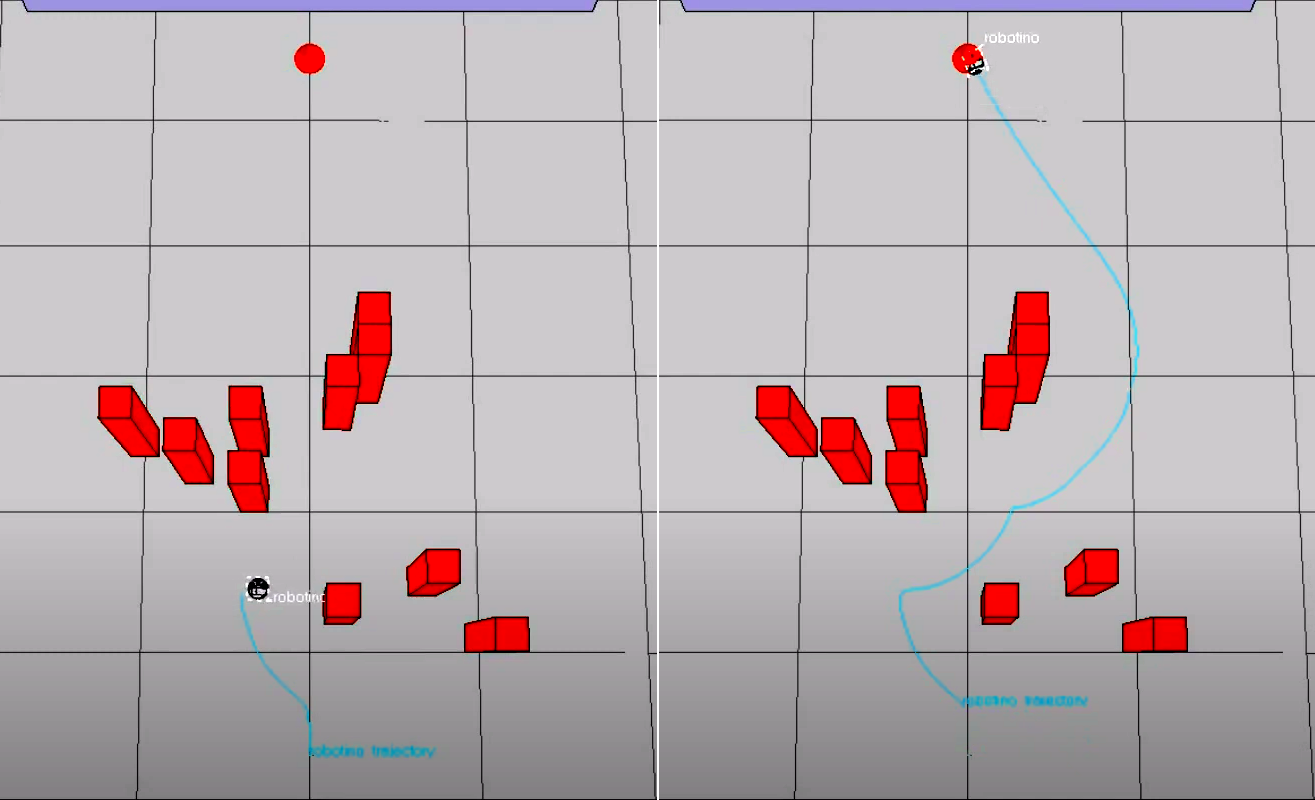}
\end{center}
\caption{Robot is passing through several stationary obstacles.}
\label{fig:mult}
\end{figure}

The resulting local planner is shown to work on several different configurations, as shown in Figures \ref{fig:vert} and \ref{fig:mult}. Since our local planner is trained on the trajectories created by SFM, we believe that the policy it learned imitates SFM's behavior. Further comparison is needed to support this claim.

\section{Limitations and Future Work}
\label{sec:future}
In this study, the preliminary results of our framework which is a hierarchical framework that is built on top of CNPs is presented. We showed that our model can generate reasonable paths at both global and local levels while avoiding obstacles. 
This work needs to be extended with a thorough statistical analysis comparing with strong baselines in successful social-aware navigation tasks. As a part of this endeavour, we plan to train our models on actual human data. Thus, we would elude the critics of SFM and prove that our model could work with real human data. Furthermore, most importantly, we plan to transfer and verify learned models in real robots.

Another direction of research to extend this work is to incorporate detectors that discover groups of people from raw sensory information. 
Human trajectory prediction can also be added to create smoother paths during navigation. For this purpose, graph neural networks \cite{tekden2020belief} that represent the world as nodes and relations between those nodes, can be employed.

CNPs have a number of drawbacks. The most important one to mention is that it cannot successfully extrapolate to the outside of the state space that it is trained on. For a mobile robot controller, this limitation is crucial since extrapolation might lead to a collision. Such cases do occur frequently when the dimensionality of the state space is high and the dataset is insufficient. We plan to learn models that detect whether the robot is trying to extrapolate and fall back to the manual controller when extrapolation occurs.


\bibliographystyle{IEEEtran} 
\bibliography{IEEEexample, IEEEabrv}

\begin{thebibliography}{10}
\providecommand{\url}[1]{#1}
\csname url@rmstyle\endcsname
\providecommand{\newblock}{\relax}
\providecommand{\bibinfo}[2]{#2}
\providecommand\BIBentrySTDinterwordspacing{\spaceskip=0pt\relax}
\providecommand\BIBentryALTinterwordstretchfactor{4}
\providecommand\BIBentryALTinterwordspacing{\spaceskip=\fontdimen2\font plus
\BIBentryALTinterwordstretchfactor\fontdimen3\font minus
  \fontdimen4\font\relax}
\providecommand\BIBforeignlanguage[2]{{%
\expandafter\ifx\csname l@#1\endcsname\relax
\typeout{** WARNING: IEEEtran.bst: No hyphenation pattern has been}%
\typeout{** loaded for the language `#1'. Using the pattern for}%
\typeout{** the default language instead.}%
\else
\language=\csname l@#1\endcsname
\fi
#2}}

\bibitem{burgard1999ExperiencesWA}
W.~Burgard, A.~Cremers, D.~Fox, D.~H{\"a}hnel, G.~Lakemeyer, D.~Schulz,
  W.~Steiner, and S.~Thrun, ``Experiences with an interactive museum tour-guide
  robot,'' \emph{Artif. Intell.}, vol. 114, pp. 3--55, 1999.

\bibitem{thrun2000probabilistic}
S.~Thrun, M.~Beetz, M.~Bennewitz, W.~Burgard, A.~B. Cremers, F.~Dellaert,
  D.~Fox, D.~Haehnel, C.~Rosenberg, N.~Roy, \emph{et~al.}, ``Probabilistic
  algorithms and the interactive museum tour-guide robot minerva,'' \emph{The
  International Journal of Robotics Research}, vol.~19, no.~11, pp. 972--999,
  2000.

\bibitem{mobot}
I.~Nourbakhsh, C.~Kunz, and T.~Willeke, ``The mobot museum robot installations:
  a five year experiment,'' in \emph{Proceedings 2003 IEEE/RSJ International
  Conference on Intelligent Robots and Systems (IROS 2003) (Cat.
  No.03CH37453)}, vol.~4, 2003, pp. 3636--3641 vol.3.

\bibitem{dwa}
D.~Fox, W.~Burgard, and S.~Thrun,
  ``\href{https://ieeexplore.ieee.org/document/580977}{The dynamic window
  approach to collision avoidance},'' \emph{IEEE Robotics Automation Magazine},
  vol.~4, no.~1, pp. 23--33, 1997.

\bibitem{nonaka2004evaluation}
S.~Nonaka, K.~Inoue, T.~Arai, and Y.~Mae,
  ``\href{https://ieeexplore.ieee.org/abstract/document/1307480}{Evaluation of
  human sense of security for coexisting robots using virtual reality. 1st
  report: evaluation of pick and place motion of humanoid robots},'' in
  \emph{IEEE International Conference on Robotics and Automation, 2004.
  Proceedings. ICRA'04. 2004}, vol.~3.\hskip 1em plus 0.5em minus 0.4em\relax
  IEEE, 2004, pp. 2770--2775.

\bibitem{helbing1995social}
D.~Helbing and P.~Molnar,
  ``\href{https://arxiv.org/abs/cond-mat/9805244}{Social force model for
  pedestrian dynamics},'' \emph{Physical review E}, vol.~51, no.~5, p. 4282,
  1995.

\bibitem{zanlungo2011social}
F.~Zanlungo, T.~Ikeda, and T.~Kanda, ``Social force model with explicit
  collision prediction,'' \emph{EPL (Europhysics Letters)}, vol.~93, no.~6, p.
  68005, 2011.

\bibitem{ferrer2013robot}
G.~Ferrer, A.~Garrell, and A.~Sanfeliu, ``Robot companion: A social-force based
  approach with human awareness-navigation in crowded environments,'' in
  \emph{2013 IEEE/RSJ International Conference on Intelligent Robots and
  Systems}.\hskip 1em plus 0.5em minus 0.4em\relax IEEE, 2013, pp. 1688--1694.

\bibitem{farina2017walking}
F.~Farina, D.~Fontanelli, A.~Garulli, A.~Giannitrapani, and D.~Prattichizzo,
  ``Walking ahead: The headed social force model,'' \emph{PloS one}, vol.~12,
  no.~1, p. e0169734, 2017.

\bibitem{kretzschmar2016socially}
``\href{https://journals.sagepub.com/doi/10.1177/0278364915619772}{Socially
  compliant mobile robot navigation via inverse reinforcement learning},
  author={Kretzschmar, Henrik and Spies, Markus and Sprunk, Christoph and
  Burgard, Wolfram},'' \emph{The International Journal of Robotics Research},
  vol.~35, no.~11, pp. 1289--1307, 2016.

\bibitem{kitani2012activity}
K.~Kitani, B.~Ziebart, J.~Bagnell, and M.~Hebert, ``Activity forecasting,''
  \emph{Computer Vision--ECCV 2012}, pp. 201--214, 2012.

\bibitem{vasquez2014inverse}
D.~Vasquez, B.~Okal, and K.~O. Arras, ``Inverse reinforcement learning
  algorithms and features for robot navigation in crowds: an experimental
  comparison,'' in \emph{2014 IEEE/RSJ International Conference on Intelligent
  Robots and Systems}.\hskip 1em plus 0.5em minus 0.4em\relax IEEE, 2014, pp.
  1341--1346.

\bibitem{kim2016socially}
B.~Kim and J.~Pineau, ``Socially adaptive path planning in human environments
  using inverse reinforcement learning,'' \emph{International Journal of Social
  Robotics}, vol.~8, no.~1, pp. 51--66, 2016.

\bibitem{wulfmeier2015maximum}
M.~Wulfmeier, P.~Ondruska, and I.~Posner, ``Maximum entropy deep inverse
  reinforcement learning,'' \emph{arXiv preprint arXiv:1507.04888}, 2015.

\bibitem{chen2017}
\BIBentryALTinterwordspacing
Y.~F. Chen, M.~Everett, M.~Liu, and J.~P. How, ``Socially aware motion planning
  with deep reinforcement learning,'' \emph{CoRR}, vol. abs/1703.08862, 2017.
  [Online]. Available: \url{http://arxiv.org/abs/1703.08862}
\BIBentrySTDinterwordspacing

\bibitem{tai2018socially}
L.~Tai, J.~Zhang, M.~Liu, and W.~Burgard, ``Socially compliant navigation
  through raw depth inputs with generative adversarial imitation learning,'' in
  \emph{2018 IEEE International Conference on Robotics and Automation
  (ICRA)}.\hskip 1em plus 0.5em minus 0.4em\relax IEEE, 2018, pp. 1111--1117.

\bibitem{gupta2018social}
A.~Gupta, J.~Johnson, L.~Fei-Fei, S.~Savarese, and A.~Alahi, ``Social gan:
  Socially acceptable trajectories with generative adversarial networks,'' in
  \emph{Proceedings of the IEEE Conference on Computer Vision and Pattern
  Recognition}, 2018, pp. 2255--2264.

\bibitem{che2020efficient}
Y.~Che, A.~M. Okamura, and D.~Sadigh, ``Efficient and trustworthy social
  navigation via explicit and implicit robot--human communication,'' \emph{IEEE
  Transactions on Robotics}, vol.~36, no.~3, pp. 692--707, 2020.

\bibitem{koren}
Y.~Koren and J.~Borenstein, ``Potential field methods and their inherent
  limitations for mobile robot navigation,'' in \emph{Proceedings. 1991 IEEE
  International Conference on Robotics and Automation}, 1991, pp. 1398--1404
  vol.2.

\bibitem{pmlr-v80-garnelo18a}
\BIBentryALTinterwordspacing
M.~Garnelo, D.~Rosenbaum, C.~Maddison, T.~Ramalho, D.~Saxton, M.~Shanahan,
  Y.~W. Teh, D.~Rezende, and S.~M.~A. Eslami,
  ``\href{https://arxiv.org/abs/1807.01613}{Conditional Neural Processes},'' in
  \emph{Proceedings of the 35th International Conference on Machine Learning},
  ser. Proceedings of Machine Learning Research, J.~Dy and A.~Krause, Eds.,
  vol.~80.\hskip 1em plus 0.5em minus 0.4em\relax PMLR, 10--15 Jul 2018, pp.
  1704--1713. [Online]. Available:
  \url{http://proceedings.mlr.press/v80/garnelo18a.html}
\BIBentrySTDinterwordspacing

\bibitem{MEYER2003283}
\BIBentryALTinterwordspacing
J.-A. Meyer and D.~Filliat,
  ``\href{https://www.sciencedirect.com/science/article/abs/pii/S138904170300007X}{Map-based
  navigation in mobile robots:: II. A review of map-learning and path-planning
  strategies},'' \emph{Cognitive Systems Research}, vol.~4, no.~4, pp.
  283--317, 2003. [Online]. Available:
  \url{https://www.sciencedirect.com/science/article/pii/S138904170300007X}
\BIBentrySTDinterwordspacing

\bibitem{astar}
S.~Kambhampati and L.~Davis,
  ``\href{https://ieeexplore.ieee.org/document/1087051}{Multiresolution path
  planning for mobile robots},'' \emph{IEEE Journal on Robotics and
  Automation}, vol.~2, no.~3, pp. 135--145, 1986.

\bibitem{giesbrecht2004global}
J.~Giesbrecht, ``\href{https://ieeexplore.ieee.org/document/6860453}{Global
  path planning for unmanned ground vehicles},'' Defence Research and
  Development Suffield (Alberta), Tech. Rep., 2004.

\bibitem{vfh}
J.~Borenstein, Y.~Koren, \emph{et~al.},
  ``\href{https://ieeexplore.ieee.org/document/88137}{The vector field
  histogram-fast obstacle avoidance for mobile robots},'' \emph{IEEE
  transactions on robotics and automation}, vol.~7, no.~3, pp. 278--288, 1991.

\bibitem{apf}
O.~Khatib, ``\href{https://ieeexplore.ieee.org/document/1087247}{Real-time
  obstacle avoidance for manipulators and mobile robots},'' in
  \emph{Proceedings. 1985 IEEE International Conference on Robotics and
  Automation}, vol.~2, 1985, pp. 500--505.

\bibitem{teb}
C.~Rösmann, F.~Hoffmann, and T.~Bertram,
  ``\href{https://ieeexplore.ieee.org/document/7331052}{Timed-Elastic-Bands for
  time-optimal point-to-point nonlinear model predictive control},'' in
  \emph{2015 European Control Conference (ECC)}, 2015, pp. 3352--3357.

\bibitem{zhu2006robot}
Q.~Zhu, Y.~Yan, and Z.~Xing,
  ``\href{https://ieeexplore.ieee.org/document/4021735}{Robot path planning
  based on artificial potential field approach with simulated annealing},'' in
  \emph{Sixth International Conference on Intelligent Systems Design and
  Applications}, vol.~2.\hskip 1em plus 0.5em minus 0.4em\relax IEEE, 2006, pp.
  622--627.

\bibitem{vadakkepat2000evolutionary}
P.~Vadakkepat, K.~C. Tan, and W.~Ming-Liang,
  ``\href{https://ieeexplore.ieee.org/document/870304}{Evolutionary artificial
  potential fields and their application in real time robot path planning},''
  in \emph{Proceedings of the 2000 congress on evolutionary computation. CEC00
  (Cat. No. 00TH8512)}, vol.~1.\hskip 1em plus 0.5em minus 0.4em\relax IEEE,
  2000, pp. 256--263.

\bibitem{cai2020mobile}
K.~Cai, C.~Wang, J.~Cheng, C.~W. De~Silva, and M.~Q.-H. Meng,
  ``\href{https://arxiv.org/abs/2006.14195}{Mobile Robot Path Planning in
  Dynamic Environments: A Survey},'' \emph{arXiv preprint arXiv:2006.14195},
  2020.

\bibitem{kruse2013human}
T.~Kruse, A.~K. Pandey, R.~Alami, and A.~Kirsch,
  ``\href{https://www.sciencedirect.com/science/article/abs/pii/S0921889013001048}{Human-aware
  robot navigation: A survey},'' \emph{Robotics and Autonomous Systems},
  vol.~61, no.~12, pp. 1726--1743, 2013.

\bibitem{seker2019conditional}
M.~Y. Seker, M.~Imre, J.~H. Piater, and E.~Ugur,
  ``\href{http://www.roboticsproceedings.org/rss15/p71.html}{Conditional Neural
  Movement Primitives.}'' in \emph{Robotics: Science and Systems}, 2019.

\bibitem{yin2019meta}
M.~Yin, G.~Tucker, M.~Zhou, S.~Levine, and C.~Finn,
  ``\href{https://arxiv.org/abs/1912.03820}{Meta-learning without
  memorization},'' \emph{arXiv preprint arXiv:1912.03820}, 2019.

\bibitem{akbulut2020adaptive}
M.~T. Akbulut, M.~Y. Seker, A.~E. Tekden, Y.~Nagai, E.~Oztop, and E.~Ugur,
  ``\href{https://arxiv.org/abs/2003.11334}{Adaptive Conditional Neural
  Movement Primitives via Representation Sharing Between Supervised and
  Reinforcement Learning},'' \emph{arXiv preprint arXiv:2003.11334}, 2020.

\bibitem{gordon2019convolutional}
J.~Gordon, W.~P. Bruinsma, A.~Y. Foong, J.~Requeima, Y.~Dubois, and R.~E.
  Turner, ``Convolutional conditional neural processes,'' \emph{arXiv preprint
  arXiv:1910.13556}, 2019.

\bibitem{coppeliaSim}
E.~Rohmer, S.~P.~N. Singh, and M.~Freese,
  ``\href{https://www.coppeliarobotics.com/coppeliaSim_v-rep_iros2013.pdf}{CoppeliaSim
  (formerly V-REP): a Versatile and Scalable Robot Simulation Framework},'' in
  \emph{Proc. of The International Conference on Intelligent Robots and Systems
  (IROS)}, 2013, www.coppeliarobotics.com.

\bibitem{robotino}
\BIBentryALTinterwordspacing
F.~Robotics, Robotino,
  ``\href{https://www.festo-didactic.com/int-en/learning-systems/factory-automation-industry-4.0/focus-trending-topics-i4.0/858/robotino-4-for-research-and-education.htm}{Robotino
  4: For research and education},'' 2020. [Online]. Available:
  \url{https://www.festo-didactic.com/int-en/learning-systems/factory-automation-industry-4.0/focus-trending-topics-i4.0/858/robotino-4-for-research-and-education.htm}
\BIBentrySTDinterwordspacing

\bibitem{tekden2020belief}
A.~E. Tekden, A.~Erdem, E.~Erdem, M.~Imre, M.~Y. Seker, and E.~Ugur,
  ``\href{https://arxiv.org/abs/1909.03785}{Belief Regulated Dual Propagation
  Nets for Learning Action Effects on Groups of Articulated Objects},'' 2020.

\end{thebibliography}

\end{document}